\documentclass{article}
\usepackage{spconf,amsmath,graphicx}

\usepackage{hyperref}
\usepackage{amsmath}
\usepackage{graphicx}	
\usepackage[utf8]{inputenc}
\usepackage{amssymb}
\usepackage{algorithm2e}
\usepackage{float}
\usepackage{caption}
\usepackage{subcaption}
\usepackage{xcolor}
\usepackage{tabularx}
\usepackage{subcaption}
\usepackage{booktabs}
\usepackage{multirow}

\newcommand{\RR}{\ensuremath{\mathbb R}}
\floatstyle{plaintop}
\restylefloat{table}
\makeatletter
\renewcommand{\@algocf@capt@plain}{above}
\newcommand{\minimize}[2]{\ensuremath{\underset{\substack{{#1}}}{\mathrm{minimize}}\;\;#2 }}

\makeatother

\title{FiGLearn: Filter and Graph Learning using Optimal Transport}
\author{}
\makeatletter
\def\@name{\emph{Matthias Minder}\quad \emph{Zahra Farsijani}\quad\emph{Dhruti Shah}  \quad \emph{Mireille El Gheche}\quad \emph{Pascal Frossard}}
\address{Ecole Polytechnique Fédérale de Lausanne (EPFL), Lausanne, Switzerland}

\usepackage[nomargin,inline,draft]{fixme}
\fxsetup{theme=color,mode=multiuser}
\FXRegisterAuthor{hpm}{ahpm}{\color{blue}matthias}
\FXRegisterAuthor{me}{ame}{\color{red}mireille}
\FXRegisterAuthor{pf}{apf}{\color{green}pascal}
\FXRegisterAuthor{gc}{agc}{\color{orange}zahra}

\begin{document}

\maketitle

\begin{abstract}
In many applications, a dataset can be considered as a set of observed signals that live on an unknown underlying graph structure. Some of these signals may be seen as white noise that has been filtered on the graph topology by a graph filter. Hence, the knowledge of the filter and the graph provides valuable information about the underlying data generation process and the complex interactions that arise in the dataset. We hence introduce a novel graph signal processing framework for jointly learning the graph and its generating filter from signal observations. We cast a new optimisation problem that minimises the Wasserstein distance between the distribution of the signal observations and the filtered signal distribution model. Our proposed method outperforms state-of-the-art graph learning frameworks on synthetic data. We then apply our method to a temperature anomaly dataset, and further show how this framework can be used to infer missing values if only very little information is available. 
\end{abstract}


\section{Introduction}

Structured data in form of graphs has become increasingly popular in scientific applications such as brain network analysis or social networks inference. Scientific research aims to use graph signal processing (GSP) or graph convolutional networks to leverage the graph topology to create better models. However, the graph topology underlying the data is often unknown, which gives rise to the problem of learning a graph from observed data. 

In the literature, various graph learning frameworks are proposed. Most frameworks rely on learning the graph representation in terms of a graph shift operator (GSO) such as the adjacency matrix or the graph Laplacian, which in turn fully characterizes the graph structure. A common assumption is that the observed signals are smooth on the unknown graph topology. Smoothness is typically related to the signal variations between nodes that are connected through an edge. In frameworks exploiting this assumption, graph learning amounts to finding the GSO that corresponds to the topology on which the signal differences across edges are minimized \cite{dong2016learning, kalofolias16, TSIPN_elgheche2019, Chierchia_neurips2019}.

Beyond the original smoothness assumptions, more recent methods assume that the observed data corresponds to some signals that were filtered on the graph with graph spectral filtering. In graph spectral filtering, a filter is applied to the eigenvalues of the GSO, while the eigenvectors remain unchanged \cite{Rue_2005_MRF,Thanou_2017}. Learning the graph thus amounts to learning the eigenvalues of the GSO before the filtering operation, while the eigenvectors can be estimated from some estimate of the covariance \cite{pasdeloup2017characterization, segarra2017network} or precision matrix \cite{ravikumar2011high, friedman2008sparse}. Assumptions have to be made about how the eigenvalues of the GSO relate to the observed eigenvalues. To this effect, authors of \cite{pasdeloup2017characterization} assume that the observations are initially independent signals which have been filtered by a diffusion process. This process is modeled by taking powers of the GSO. Meanwhile, authors of \cite{segarra2017network} assume that the optimal GSO is sparse. 

However, in case of few data being available, the eigenvectors of the GSO can only be poorly approximated using the data. This motivates methods that jointly learns eigenvectors and eigenvalues. Authors of \cite{thanou2017learning} assume that the observed signals can be decomposed into a sum of heat kernel signals with different diffusion times. They simultaneously learn the diffusion time, the localization of the heat signals and the GSO. In \cite{egilmez2017graph}, the authors propose a framework in which the GSO can be seen as the maximum likelihood estimator. This framework is extended in \cite{egilmez2018graph} where the authors propose an algorithm to simultaneously learn the GSO and the graph filter. The graph filter is assumed to stem from a filter family with one learnable parameter. 

In all those methods, strong assumptions about the nature of the filter must be made. The most flexibility is provided in \cite{egilmez2018graph} and \cite{thanou2017learning} where some aspects of the filter are learned. Nevertheless, they make strong assumptions about the filter: \cite{egilmez2018graph} assumes that the filter is known up to some parameter, whereas \cite{thanou2017learning} assumes a sum of heat kernels. Since in many applications those assumptions are too strong, we introduce a framework that models the filter as neural network, and thus constrains the filter only to be continuous. We further propose a new loss function that minimises the distance between the observed signal distribution, and the filtered signal distribution model. This distance is captured with optimal transport, which permits to develop a notion of distance between graphs that is better suited to capture structural properties of the data \cite{maretic2019got}. We solve this problem by alternative optimisation and accelerated gradient descent, and show that our learning framework achieves better results than baseline algorithms in learning tasks, but also in data completion applications. 
\section{Graph and filter estimation}
\label{sec:graph_filter_estimation}

\subsection{Graph signal processing}
\label{sub:gsp}
Within this work, we consider an undirected, unweighted graph $\mathcal{G}=(V,E)$ with a set $V$ of $N$ vertices and a set $E$ of edges. The adjacency matrix is denoted by $W \in \mathbb{R}^{N \times N}$. The degree of a vertex $i\in V$, denoted by $d(i)$, is the number of edges incident to $i$ in the graph $\mathcal{G}$. The degree matrix $D\in \RR^{N\times N}$ is defined as $D = {\rm Diag}\left(d(1),\cdots,d(N)\right)$.
Based on $W$ and $D$, the Laplacian matrix of $\mathcal{G}$ is $L = D-W.$
As the graph Laplacian is a real symmetric matrix, it has a complete set of orthonormal eigenvectors $U=[U_0, \cdots, U_{N-1}]$ and a corresponding set of non-negative eigenvalues $\Lambda= {\rm Diag} [\lambda_0, \cdots, \lambda_{N-1}]$. 

In addition, we consider some attributes modeled as features on the graph vertices. Assuming that each node $i$ in $V$ is associated to a feature vector $x\in\RR^M$, the graph signal is represented as $X\in \mathbb{R}^{N\times M}$. Following \cite{Rue_2005_MRF,Thanou_2017}, we will consider a graph signal $x$ on a given graph $\mathcal{G}$ as an initial signal $x^0 \sim \mathcal{N}(0,{\rm Id})$ that is filtered with the graph filter $h(\cdot)$ to yield 
\begin{equation}
    x = h(L) \, x^0.
\end{equation}
Therefore, $x$ follows a normal distribution given by\footnote{We denote by $h$ both the function $h: \RR \rightarrow \RR$ and its matrix counterpart $h : \RR^{N \times N} \rightarrow \RR^{N \times N}$ acting on the matrix’s eigenvalues.}
\begin{align}
\label{eq:gaussain_graph}
    \nu^{\mathcal{G}} = \mathcal{N}\big(0, h^2(L)\big).
\end{align}

The graph filter $h(\cdot)$ carries the spatial and the spectral characteristics of the graph and is defined as follows:
\begin{equation}
\label{eq:eig_fil}
    h(L) = U \textbf{H}(\Lambda) U^\top,
\end{equation}
with a diagonal matrix
\begin{equation}
\label{eq:eig_fil_lambda}
    \textbf{H}_{i,j} (\Lambda)= \begin{cases} h(\lambda_i) \quad & \textup{if $i=j$}, \\
0 \quad & \textup{otherwise}.
\end{cases}
\end{equation}

\subsection{Distribution comparisons}
\label{sub:ot}

Graph signal distributions can take different forms, depending on the graph filter and on the actual graph that underlies the graph signals. In order to compare distributions, we need tools that are able to properly handle the structure of the data, such as those based on the optimal transport framework. Optimal transport (OT) is a fundamental notion in probability theory \cite{villani2008optimal,simou2019_graphbary,Ortiz2020_OTDA} that defines a notion of distance between probability distributions by quantifying the optimal displacement of mass according to a so-called ground cost associated to the geometry of a supporting space.
In particular, given two normal distributions  $\nu_1 = \mathcal{N}(m_1, \Sigma_1)$ and $\nu_2 = \mathcal{N}(m_2, \Sigma_2)$, one can compute the distance between the probability measures through the 2-Wasserstein distance. More precisely, the 2-Wasserstein distance corresponds to the minimal ``effort'' required to transport $\nu_1$ to $\nu_2$ with respect to the Euclidean norm \cite{Monge_1781}, that is
\begin{equation}\label{eq:wass}
W_2^2\big(\nu_1, \nu_2\big) = \inf_{T_{\#}\nu_1 = \nu_2} {\int_{\mathcal{X}}  \|x - T(x)\|^2 \, d\nu_1(x)},
\end{equation}
where $T_{\#}\nu_1$ denotes the push-forward of $\nu_1$ by the transport map $T\colon \mathcal{X} \to \mathcal{X}$ defined on a metric space $\mathcal{X}$. For normal distributions such as $\nu_1$ and $\nu_2$, the 2-Wasserstein distance can be explicitly written in terms of their means and covariance matrices \cite{Takatsu2011,maretic2019got}, yielding\footnote{$\|\cdot\|_F$ is the Frobenius norm.}
\begin{equation}
\label{eq:wasserstein_def}
    \mathcal{W}_2^2(\nu_1, \nu_2) = \|m_1 - m_2\|_2^2 + \| \Sigma_1^{1/2} - \Sigma_2^{1/2} \|_F^2.
\end{equation}

\subsection{Problem formulation}

We now formulate a problem that learns the unknown filter $h$ and graph $L$ that best explains the observed signals. In other words, we look for the best combination of $h$ and $L$ such that $\nu^\mathcal{G}$ is close in distribution to the observed signal. 

The observed graph signals are denoted by $\bar{X} \in \RR^{M\times N}$. We can estimate the sample mean $\hat{m}$ and covariance matrix $\hat{C}$ from the signals $\bar{X}$, yielding the estimated graph signal distribution 
\begin{align}
    \hat{\nu} = \mathcal{N} (\hat{m}, \hat{C}).
\end{align}

Following the probabilistic model introduced in Section \ref{sub:gsp}, the observed signals in $\bar{X}$ should follow a Gaussian distribution driven by a graph filter $h$ and a graph $L$. In other words, we assume that $\forall i \in \{1,\cdots, N\}$, $\bar{x}_i$ follows a Gaussian distribution
\begin{align}
    \bar{x}_i \sim \nu^{\mathcal{G}} = \mathcal{N}(0, h^2(L)).
\end{align}

We propose to recover $L$ and $h(\cdot)$ from $\bar{X}$ by minimizing the Wasserstein distance in Eq. \eqref{eq:wasserstein_def} between the theoretical distribution $\nu^{\mathcal{G}}$ and the estimated distribution $\hat{\nu}$, namely
\begin{equation}
\label{eq:wasserstein_L}
    \minimize{L\in\mathbb{R}^{N\times N}, h}{\| \hat{C}^{\frac{1}{2}} - h(L) \|_F^2 + \|\hat{m}\|_2^2}, \quad {\rm s.t.} \quad L \in \mathcal{C},
\end{equation}
where $\mathcal{C}$ is the space of valid combinatorial graph Laplacians defined as follows
\begin{equation}
    \mathcal{C} = \{L\in\mathbb{R}^{N\times N} \,|\, \; L_{ij}=L_{ji}\leq 0, L_{ii}=-\sum_{j\neq i} L_{ij}\}.
\end{equation}
The main difficulty of this optimization problem arises from the non-convexity of the Wasserstein distance with respect to $L$ and $h$. Hence, we divide the optimization problem into two subtasks, namely learning i) the filter $h$ and ii) the graph Laplacian $L$. When neither the filter nor the Laplacian are known, we alternate between learning the graph and the filter.

\section{Algorithms} \label{sec:algorithms}

In this section, we propose an alternating optimization algorithm to solve Problem \eqref{eq:wasserstein_L}. Both optimization problems are fully differentiable, and thus can be solved via accelerated gradient descent algorithm (Adam) \cite{Reddi2018}.\footnote{The function gradients are obtained automatically using the auto-gradient functionality of PyTorch \cite{NEURIPS2019_9015}.} 

\paragraph*{Filter learning}

Given the graph $L$, we aim to estimate the filter $h$ that describes the data generation process well. This amounts to the following minimization:
\begin{equation}
\label{eq:filter_learning}
   \minimize{h}{\| \hat{C}^{\frac{1}{2}} - h(L) \|_F^2 + \|\hat{m}\|_2^2}.
\end{equation}
Note that when we apply a filter over a graph, only the eigenvalues of $L$ are taken into account \big(Eqs. \eqref{eq:eig_fil} and \eqref{eq:eig_fil_lambda}\big). As a consequence, the estimated filter is thus only fitted at the eigenvalue positions. We will assume that the filter is continuous between the observed eigenvalues.

In order to solve the minimization problem formulated in equation \eqref{eq:filter_learning}, the filter $h$ consists of a fully connected feed-forward neural network with 5 hidden layers of 30 neurons each, yielding the output $f_\theta(L)$, where $\theta$ denotes the parameter vector of the neural network. We apply the softplus function to the output of the network in order to assure that the final filter $h$ maps all values to positive real numbers. The expression for the filter hence becomes $h(x) = \ln\big(1+\exp\left(f_\theta(L)\right)\big)$ and Eq. \eqref{eq:filter_learning} can be rewritten as
\begin{equation}
\label{eq:filter_learning_theta}
    \minimize{\theta}{ \| \hat{C}^{1/2} - \ln\big(1 + \exp\left(f_{\theta}(L)\right)\big) \|_F^2 + \|\hat{m}\|_2^2.}
\end{equation}

\paragraph*{Graph learning}

To facilitate the optimization defined in equation \eqref{eq:wasserstein_L} with respect to the graph structure, we seek for a valid adjacency matrix\footnote{$\mathcal{W} = \{W\in\mathbb{R}^{N\times N} \,|\, \; W=W^\top, {\rm diag}W=0, W_{ij} \in \{0, 1\}\}.$} instead of a Laplacian. Moreover, we can exploit the symmetry of the adjacency matrix by only optimizing the upper triangular part $w$. If we denote as $\mathcal{L}\colon \RR^{(N-1)N/2} \rightarrow \RR^{N\times N}$ the linear operator that converts the upper-triangular part $w$ of an adjacency matrix into the corresponding Laplacian matrix, we can reformulate the problem as
\begin{equation}
   \minimize{w\in\{0,1\}^{(N-1)N/2}}{\| \hat{C}^{\frac{1}{2}} - h(\mathcal{L}w) \|_F^2 + \|\hat{m}\|_2^2}.
\end{equation}
To solve our problem with gradient descent, we further relax the binary constraint into the unitary interval $w\in [0,1]^{(N-1)N/2}$. To avoid a projection step, we express $w$ in terms of some $z \in \mathbb{R}^{(N-1)N/2}$ as $w=1/(1 + exp\{-z\})$. We disregard the mean $\hat{m}$ as it is constant with respect to the parameters, to finally obtain\footnote{The values for $z$ are initialized uniformly at random in $[-0.5, 0.5]$.}
\begin{equation}
   \minimize{z\in\mathbb{R}^{(N-1)N/2}}{\Big\| \hat{C}^{\frac{1}{2}} - h\left(\mathcal{L}\left(\frac{1}{1 + {\rm exp}(-z)}\right)\right) \Big\|_F^2}.
\end{equation}

\begin{table}
\centering
{
\caption{Graph learning performance.\label{tbl:performance}}
\begin{tabular}{lllll}
\toprule
Algorithm & F1-score  & Precision & Recall & Accuracy \\ 
\midrule
\multicolumn{1}{l}{$h_{\rm heat}$} & & & & \vspace{0.5em} \\
\textbf{FiGLearn} & 0.8791 & 0.8952 & 0.8637 & 0.9454 \\ 
\textbf{GSI} & 0.7903 & 0.7450  & 0.9147 & 0.8621 \\ 
 \textbf{CGL}& 0.4697 & 0.3107 & 1.0 & 0.3676 \\ 
\midrule
\multicolumn{1}{l}{$h_{\rm normal}$} & & & &\vspace{0.5em} \\
\textbf{FiGLearn} & 0.6660 & 0.8352 & 0.5649 & 0.8254 \\ 
\textbf{GSI} & 0.6137 & 0.5570 & 0.7271 & 0.7199 \\ 
\textbf{CGL}& 0.6356 & 0.4736 & 0.9739 & 0.6865 \\ 
\midrule
\multicolumn{1}{l}{$h_{\rm high-pass}$} & & & & \vspace{0.5em} \\
\textbf{FiGLearn} & 0.4960 & 0.5742 & 0.4514 & 0.7300 \\ 
\textbf{GSI} & 0.4878 & 0.3606 & 0.9030& 0.4297 \\ 
 \textbf{CGL}& 0.2551 & 0.2051 & 0.3455 & 0.3711 \\ 
\bottomrule
\end{tabular}}
\end{table}


\section{Experiments} \label{sec:experiments}
The numerical analysis is decomposed in two parts. Firstly, we assess the performance achieved when solving Problem Eq.~\eqref{eq:wasserstein_L} on synthetic data. Secondly, we demonstrate the application of our framework to missing data inference in a temperature data set. 

\begin{figure*}
     \centering
         \includegraphics[width=1\textwidth]{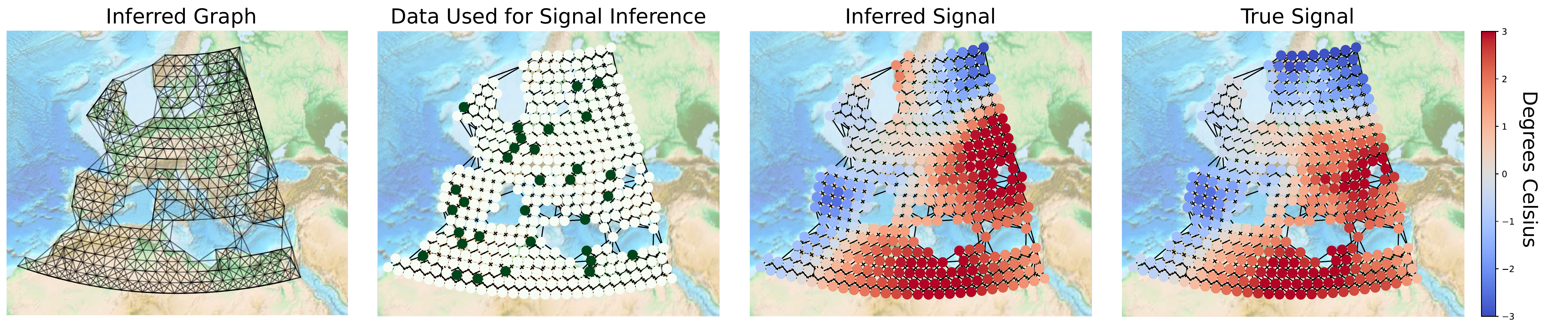}
        \caption{Results of missing value inference. From the \textit{left} to the \textit{right}: the inferred graph, the nodes for which data are available are marked in green, the inferred signal, the true signal.}
        \label{fig:res_missing}
\end{figure*}

\paragraph*{Synthetic data}

To compare the performance of our algorithm to other algorithms, we randomly generate 60 graphs from stochastic block model partitioned into graphs with 30, 50, and 70 nodes  with probability of an edge between clusters of $0.1$, and a probability of an edge within clusters of $0.3$, $0.5$ and $0.7$, respectively. On each graph, we generate $500$ signals by filtering white noise with different filters as in Eq. \eqref{eq:gaussain_graph}, in particular the following:
\begin{align} 
    h_{\rm normal}(x) & = \begin{cases}
    \frac{1}{\sqrt{x}},& \text{if } x > 0\\
    0,              & \text{if } x=0,
    \end{cases} \label{eq:kernels_normal}\\
    h_{\rm heat}(x) & = \exp\left(-0.1\,x\right), \label{eq:kernels_heat}\\
    h_{\rm high-pass}(x) &= \frac{0.1\,x}{1+0.1\,x}. \label{eq:kernels_high}
\end{align}
For each combination of stochastic block model and graph filter, we repeat the experiment 20 times. We assess the method's performance in a single experiment by calculating the F1-score between the true and the learned Laplacian matrices. We compare our method with the current state-of-the-art methods GSI \cite{egilmez2018graph} and CGL \cite{egilmez2017graph}. Note that CGL is merely a graph learning algorithm and not capable of learning graph filter and GSI requires prior information about the type of the one-to-one filter (e.g., normal, heat, or high-pass) which we feed into the model. Hence, both CGL and GSI are less powerful than FiGLearn which learns the filter without having any prior knowledge about the filter. The results averaged over all 60 graphs are reported in Table \ref{tbl:performance}. 

According to the results, our algorithm, FiGLearn, surpasses the other algorithms, namely, GSI and CGL, in performance in terms of our evaluation metrics. FiGLearn achieves the highest F1-score for all the three filters considered, which means that the graph learned by FiGLearn is the closest to the ground truth in topology.
\paragraph*{Inference of Missing Temperature Data}

We demonstrate an application of our approach in which we have a good estimate for the covariance matrix, but have some observations with many missing values. We emulate this on a temperature anomaly dataset \cite{BerkeleyEarth} with 397 measure points in Europe and North Africa. We split the data into a complete set (95\%, 2042 observations) that will be used to calculate the covariance matrix and infer the graph, and an incomplete set (5\%, 122 observations) for which we will infer the missing values. In the incomplete set, we only retain 10\% of the data entries that are chosen at random, all other data will be considered as missing. The task thus amounts to inferring missing values when only very little information is available.  


To infer missing values, we first learn a graph and its filter on the complete set using our method described in Section \ref{sec:algorithms}. Visual inspection of the graph (Fig. \ref{fig:res_missing}) shows that the graph connects adjacent quadrants. Some meteorological features are captured as well: There is no edge crossing the alps, capturing that the climate in the north and south of the alps is very different. 

Using the learned graph Laplacian $L$ and filter $h$, we can now infer the missing values as follows: Let $y\in\RR^N$ be the temperature signal for which the values $\{y_i\}_{i\in \mathcal{I}}$ are available ($\forall i\notin \mathcal{I}, \,y_i = \texttt{NA}$). We infer missing values by finding the generating signal $x$ that, when filtered on the graph with the learned filter, is close to $y$ in terms of the mean absolute error at the available positions $\mathcal{I}$. Formally, we solve the following optimization problem
\begin{equation}
    \minimize{x \in \mathbb{R}^N}{\sum_{i \in \mathcal{I}}  \big|\big(h(L)x\big)_i - y_i\big|^2}.
\end{equation}
$x$ is initialized as white noise, and the problem is solved with Adam \cite{Reddi2018}. 

On the data in the incomplete set, this inference method achieves an MSE of 0.195 in average, and a standard deviation of 0.136. Figure \ref{fig:res_missing} shows the inference with the highest MSE. We observe that the method strongly underestimates temperature in Norway, where it has no information that would suggest the unusually high temperatures there. It further overestimates temperatures north of the black sea, but other than that still is able to capture the overall dynamics very well.  

This demonstrates that our graph inference method can infer a meaningful graph of medium size (around 400 nodes, computation takes a few minutes). Moreover, it shows how the knowledge of the filter can be leveraged to understand how the nodes interact, which in this case was used to infer missing data from very few observations. 


\section{Conclusion} \label{sec:conclusion}

We introduced a new framework\footnote{The code is available online: github.com/mattminder/FiGLearn} for graph inference, which outperforms the current state-of-the-art methods. The better performance may be explained by the fact that our approach jointly learns the graph and the filter generating the signal, thus taking advantage of important information about the way the data is generated. We demonstrated how this knowledge can be used by accurately inferring missing values when only very little data is available.


Some applications may require weights to be outside of the interval between 0 and 1. Our framework can easily be extended to weighted graphs, which will open new challenges concerning the filter priors in order to have a better estimate with respect to the real settings. 

Another perspective is to explore dynamic graphs, where timing information is critical, e.g., instant message networks or financial transaction graphs. It will be interesting to explore the task of incorporating timing information about edges and to develop a complete approach to solve (spatio-)temporal graph and filter estimation.




\bibliographystyle{IEEEbib}
\bibliography{strings,bibfile}

\begin{thebibliography}{10}

\bibitem{dong2016learning}
X.~Dong, D.~Thanou, P.~Frossard, and P.~Vandergheynst,
\newblock ``Learning laplacian matrix in smooth graph signal representations,''
\newblock {\em IEEE Transactions on Signal Processing}, vol. 64, no. 23, pp.
  6160--6173, 2016.

\bibitem{kalofolias16}
V.~Kalofolias,
\newblock ``How to learn a graph from smooth signals,''
\newblock in {\em Proceedings of the 19th International Conference on
  Artificial Intelligence and Statistics}, Arthur Gretton and Christian~C.
  Robert, Eds., Cadiz, Spain, 09--11 May 2016, vol.~51 of {\em Proceedings of
  Machine Learning Research}, pp. 920--929, PMLR.

\bibitem{TSIPN_elgheche2019}
M.~El~Gheche, G.~Chierchia, and P.~Frossard,
\newblock ``Orthonet: Multilayer network data clustering,''
\newblock {\em IEEE Transactions on Signal and Information Processing over
  Networks}, vol. 6, no. 1, pp. 13--23, Dec. 2020.

\bibitem{Chierchia_neurips2019}
G.~Chierchia and B.~Perret,
\newblock ``Ultrametric fitting by gradient descent,''
\newblock in {\em Advances in Neural Information Processing Systems 32},
  H.~Wallach, H.~Larochelle, A.~Beygelzimer, F.~d'Alch\'{e} Buc, E.~Fox, and
  R.~Garnett, Eds., pp. 3175--3186. Curran Associates, Inc., 2019.

\bibitem{Rue_2005_MRF}
H.~Rue and L.~Held,
\newblock {\em Gaussian Markov Random Fields: Theory And Applications
  (Monographs on Statistics and Applied Probability)},
\newblock Chapman \& Hall CRC, 2005.

\bibitem{Thanou_2017}
D.~{Thanou}, X.~{Dong}, D.~{Kressner}, and P.~{Frossard},
\newblock ``Learning heat diffusion graphs,''
\newblock {\em IEEE Transactions on Signal and Information Processing over
  Networks}, vol. 3, no. 3, pp. 484--499, 2017.

\bibitem{pasdeloup2017characterization}
B.~Pasdeloup, V.~Gripon, G.~Mercier, D.~Pastor, and M.~G. Rabbat,
\newblock ``Characterization and inference of graph diffusion processes from
  observations of stationary signals,''
\newblock {\em IEEE transactions on Signal and Information Processing over
  Networks}, vol. 4, no. 3, pp. 481--496, 2017.

\bibitem{segarra2017network}
S.~Segarra, A.~G Marques, G.~Mateos, and A.~Ribeiro,
\newblock ``Network topology inference from spectral templates,''
\newblock {\em IEEE Transactions on Signal and Information Processing over
  Networks}, vol. 3, no. 3, pp. 467--483, 2017.

\bibitem{ravikumar2011high}
P.~Ravikumar, M.~J. Wainwright, G.~Raskutti, B.~Yu, et~al.,
\newblock ``High-dimensional covariance estimation by minimizing l1-penalized
  log-determinant divergence,''
\newblock {\em Electronic Journal of Statistics}, vol. 5, pp. 935--980, 2011.

\bibitem{friedman2008sparse}
J.~Friedman, T.~Hastie, and R.~Tibshirani,
\newblock ``Sparse inverse covariance estimation with the graphical lasso,''
\newblock {\em Biostatistics}, vol. 9, no. 3, pp. 432--441, 2008.

\bibitem{thanou2017learning}
D.~Thanou, X.~Dong, D.~Kressner, and P.~Frossard,
\newblock ``Learning heat diffusion graphs,''
\newblock {\em IEEE Transactions on Signal and Information Processing over
  Networks}, vol. 3, no. 3, pp. 484--499, 2017.

\bibitem{egilmez2017graph}
H.~E. Egilmez, E.~Pavez, and A.~Ortega,
\newblock ``Graph learning from data under laplacian and structural
  constraints,''
\newblock {\em IEEE Journal of Selected Topics in Signal Processing}, vol. 11,
  no. 6, pp. 825--841, 2017.

\bibitem{egilmez2018graph}
H.~E. Egilmez, E.~Pavez, and A.~Ortega,
\newblock ``Graph learning from filtered signals: Graph system and diffusion
  kernel identification,''
\newblock {\em IEEE Transactions on Signal and Information Processing over
  Networks}, vol. 5, no. 2, pp. 360--374, 2018.

\bibitem{maretic2019got}
H.~P. Maretic, M.~El~Gheche, G.~Chierchia, and P.~Frossard,
\newblock ``Got: An optimal transport framework for graph comparison,''
\newblock in {\em Advances in Neural Information Processing Systems}, 2019, pp.
  13876--13887.

\bibitem{villani2008optimal}
C.~Villani,
\newblock {\em Optimal transport: old and new}, vol. 338,
\newblock Springer Science \& Business Media, 2008.

\bibitem{simou2019_graphbary}
E.~{Simou} and P.~{Frossard},
\newblock ``Graph signal representation with wasserstein barycenters,''
\newblock in {\em IEEE International Conference on Acoustics, Speech and Signal
  Processing}, 2019, pp. 5386--5390.

\bibitem{Ortiz2020_OTDA}
G.~{Ortiz-Jiménez}, M.~{El Gheche}, E.~{Simou}, H.~P. {Maretić}, and
  P.~{Frossard},
\newblock ``Forward-backward splitting for optimal transport based problems,''
\newblock in {\em IEEE International Conference on Acoustics, Speech and Signal
  Processing}, 2020, pp. 5405--5409.

\bibitem{Monge_1781}
M.~Monge,
\newblock {\em M\'emoire sur la th\'eorie des d\'eblais et des remblais},
\newblock De l'Imprimerie Royale, 1781.

\bibitem{Takatsu2011}
A.~Takatsu,
\newblock ``Wasserstein geometry of gaussian measures,''
\newblock {\em Osaka Journal of Mathematics}, vol. 48, no. 4, pp. 1005--1026,
  2011.

\bibitem{Reddi2018}
S.~J. Reddi, S.~Kale, and S.~Kumar,
\newblock ``On the convergence of adam and beyond,''
\newblock in {\em International Conference on Learning Representations},
  Vancouver, Canada, May 2018.

\bibitem{NEURIPS2019_9015}
A.~Paszke, S.~Gross, F.~Massa, A.~Lerer, J.~Bradbury, G.~Chanan, T.~Killeen,
  Z.~Lin, N.~Gimelshein, L.~Antiga, A.~Desmaison, A.~Kopf, E.~Yang, Z.~DeVito,
  M.~Raison, A.~Tejani, S~Chilamkurthy, B.~Steiner, L.~Fang, J.~Bai, and
  S.~Chintala,
\newblock ``Pytorch: An imperative style, high-performance deep learning
  library,''
\newblock in {\em Advances in Neural Information Processing Systems 32},
  H.~Wallach, H.~Larochelle, A.~Beygelzimer, F.~d'Alch\'{e} Buc, E.~Fox, and
  R.~Garnett, Eds., pp. 8024--8035. Curran Associates, Inc., 2019.

\bibitem{BerkeleyEarth}
``Average monthly temperature anomaly data on an equal space grid,''
  \url{berkeleyearth.org/data-new/},
\newblock Accessed: 2020-10-19.

\end{thebibliography}


\end{document}